\documentclass{article}
\usepackage{ijcai24}

\usepackage{times}
\usepackage{arydshln}
\usepackage{multirow}
\usepackage{soul}
\usepackage{dsfont}
\usepackage{url}
\usepackage[utf8]{inputenc}
\usepackage[small]{caption}
\usepackage{graphicx}
\usepackage{amsmath}
\usepackage{amsthm}
\usepackage{booktabs}
\usepackage{algorithm}
\usepackage{algorithmic}
\usepackage[switch]{lineno}
\usepackage{newclude}
\usepackage{amsfonts,amssymb}
\usepackage[hidelinks]{hyperref} 
\makeatletter
\renewcommand{\maketag@@@}[1]{\hbox{\m@th\normalsize\normalfont#1}}%
\makeatother

\usepackage{float}
\usepackage{subfigure}

\def\onedot{\ifx\@let@token.\else.\null\fi\xspace}

\makeatother

\newcommand\etal{\emph{et al.}}

\title{Personalized Heart Disease Detection via ECG Digital Twin Generation}

\author{
Yaojun Hu$^1$
\and
Jintai Chen$^{2,}$\thanks{Corresponding Author. E-mail: \url{jtchen721@gmail.com}}
\and
Lianting Hu$^{3,4,5}$\and
Dantong Li$^{3,4,5}$\and
Jiahuan Yan$^1$\and \\
Haochao Ying$^6$\and
Huiying Liang$^{3,4,5}$\and
Jian Wu$^{6,7}$
\affiliations
{\small
$^1$College of Computer Science and Technology, Zhejiang University\\
$^2$Computer Science Department, University of Illinois at Urbana-Champaign\\
$^3$Medical Big Data Center, Guangdong Provincial People’s Hospital, Southern Medical University\\
$^4$Guangdong Cardiovascular Institute, Guangdong Provincial People’s Hospital, Guangdong Academy of Medical Sciences\\
$^5$Guangdong Provincial Key Laboratory of Artificial Intelligence in Medical Image Analysis and Application, Guangdong Provincial People’s Hospital (Guangdong Academy of Medical Sciences)\\
$^6$School of Public Health, Zhejiang University\\
$^7$State Key Laboratory of Transvascular Implantation Devices of The Second Affiliated Hospital, Zhejiang University School of Medicine}
\emails
}

\begin{document}

\maketitle
\begin{abstract}
Heart diseases rank among the leading causes of global mortality, demonstrating a crucial need for early diagnosis and intervention. Most traditional electrocardiogram (ECG) based automated diagnosis methods are trained at population level, neglecting the customization of personalized ECGs to enhance individual healthcare management. A potential solution to address this limitation is to employ digital twins to simulate symptoms of diseases in real patients. In this paper, we present an innovative \textit{prospective learning} approach for personalized heart disease detection, which generates digital twins of healthy individuals' anomalous ECGs and enhances the model sensitivity to the personalized symptoms. In our approach, a \textit{vector quantized feature separator} is proposed to locate and isolate the disease symptom and normal segments in ECG signals with ECG report guidance. Thus, the ECG digital twins can simulate specific heart diseases used to train a personalized heart disease detection model. Experiments demonstrate that our approach not only excels in generating high-fidelity ECG signals but also improves personalized heart disease detection. Moreover, our approach ensures robust privacy protection, safeguarding patient data in model development. The code can be found at \url{https://github.com/huyjj/LAVQ-Editor}.

\end{abstract}

\section{Introduction}
Heart disease stands as a leading cause of global mortality~\cite{roth2018global}. 
In clinical practice, the electrocardiogram (ECG) is the most routinely-used tool for heart disease detection due to its affordability, convenience, and non-invasiveness. However, the lengthy ECG signals (e.g., records that may span several hours from Holter monitors) pose challenges for manual inspection, and cutting-edge deep learning techniques~\cite{li2023mets,chen2020flow,chen2024congenital} are widely employed to automatically find anomalous signals. Nevertheless, the application of ECG signals in deep learning is constrained by high annotation costs and data privacy concerns. In addition, given the variability of disease symptoms among individual patients, current data-driven deep learning approaches face challenges in ensuring the reliability across diverse patient profiles.

An obvious solution to the data dilemma is to generate more data by deep generative networks. Deep generative networks such as generative adversarial networks (GANs)~\cite{goodfellow2014gan} and variational autoencoders (VAEs)~\cite{kingma2013vae} have been used to synthetic ECGs~\cite{hossain2021ecgadvgan, Chen2023megan}. Nevertheless, most prior researches~\cite{Golany2020simgan,Chen2023megan,chen2021electrocardio} have predominantly concentrated on generating ECGs following population distributions, largely neglecting the creation of tailored ECGs for individual patients. 
Such limitation suggests that the generated ECGs are more aptly utilized as supplementary data for training models rather than for direct application in personalized patient care within clinical practice.
Digital twins are virtual constructs designed to mirror real-world objects, allowing for rapid and non-invasive simulation of disease progression and treatment trials~\cite{das2023twin}. It also involves preserving unique patient information to ensure that subsequent simulations and predictive modeling are meaningful.
To specify, the primary goal of this paper is to generate personalized ECG digital twins, thereby facilitating the individualized management of heart conditions.

We create personalized ECG digital twins by editing an individual's ECG signals to simulate the heart disease symptoms prospectively for heart disease detection improvement. 
Our purpose is to ensure that the created ECG digital twins can assist clinicians and automated diagnosis machine learning approaches in gaining prospective insight into individual symptoms, thereby enhancing the quality of personalized diagnosis.
The representations of most heart diseases in ECG signals typically exhibit locality, attributed to the different signal wave segments reflecting conditions at specific parts on the heart~\cite{Chen2023megan}. Hence, when simulating a target disease, our approach separates the ECG features into normal segments and disease-indicative segments guided by the textual description of the disease. The disease-indicative components are then edited to align with the features of the target disease. In this paper, we define ``disease-indicative feature'' as the ECG features wherein abnormalities may manifest, serving as the basis upon which clinicians rely for diagnosing the heart disease. Notably, distinct heart diseases may exhibit on different disease-indicative segments. 

Our proposed model comprises three components: the VQ-Separator for text-guided separation of different segments of the ECG, a generator employing a feature mapper to incorporate disease information into the ECG signals, and a discriminator to discern authenticity. 
In the training process, we employ ECG-text pairs from a \textit{pre-diagnosis patient} whose ECG digital twins are to be constructed (potentially healthy or otherwise) and a \textit{reference patient} already diagnosed with a specific disease. According to the guidance of the heart disease descriptive text, VQ-Separator picks out the disease-indicative features from the personal normal features. 
Subsequently, we leverage the normal feature of the \textit{pre-diagnosis patient} and reference the disease-indicative feature from the \textit{reference patient} to adjust the ECGs of the \textit{pre-diagnosis patient}, aligning the disease condition with that of the \textit{reference patient}. 
In addition, to preserve personalized characteristics in the ECG signals, the generator not only learns to generate edited ECGs but also possesses the capability to reconstruct ECGs with their own disease-indicative features. The discriminator is trained to discern both the validity and whether the edited ECGs have integrated the desired disease features. Our main contributions are summarized as follows:
\begin{itemize}
    \item We introduce a novel \textit{prospective learning} concept for personalized heart detection by creating personalized ECG digital twins that simulate heart disease symptoms. This method provides prospective information to enhance the cognition of heart diseases, thereby improving subsequent detection performance.
    \item We propose a location-aware model called LAVQ-Editor for personalized ECG digital twins generation, which localizes and separates personal normal features and disease-indicative features guided by textual disease descriptions. This is achieved through the novel VQ-Separator that operates with a Disease Embedding Codebook to precisely model the disease features.
    \item Empirically, our controlled experiments have demonstrated that patients who provided prospective cognition to the heart disease detection model via ECG digital twins achieved significantly better diagnostic accuracy than those who did not. Furthermore, our experiments also validate that our methodology not only efficiently leverages patient information but also enhances patient privacy to a greater extent.
    
\end{itemize}

\section{Related Work}
\subsection{Generative Adversarial Networks}
Currently, generative adversarial networks have made great achievements in the areas of image synthesis~\cite{Karras2019stylegan}, text  generation~\cite{yu2017seqgan} and scientific discovery~\cite{repecka2021expanding}. In addition, GANs have been instrumental in enabling interactive image editing where users can provide input to guide the generation process. For instance, GauGAN~\cite{park2019Gaugan} allowed users to create complex scenes by painting a segmentation map that the network converts into a photorealistic image, effectively turning sketches into detailed landscapes. Patashnik \etal~\cite{patashnik2021styleclip} introduced StyleCLIP, which utilizes the capabilities of StyleGAN and CLIP to manipulate images based on textual descriptions. The method is known for its intuitive nature, allowing non-experts to make significant, detailed changes to images using simple text prompts. Hertz \etal~\cite{hertz2022prompt} presented a novel approach that refines the interaction between text prompts and the generated images using a cross-attention within transformer~\cite{vaswani2017attention}. This advancement allows for more precise and localized control over the image editing process, facilitating detailed adjustments without extensive collateral impacts on the image. Similarly, our method separates disease-indicative and normal features in ECGs through text-guidance and cross-attention, and then edits the ECG by fusing the target disease-indicative features.
\subsection{Vector Quantization based Methods}
The Vector Quantized Variational Autoencoder (VQ-VAE)~\cite{van2017vqvae} is a significant advancement in the field of generative models, blending the concepts of auto encoding and vector quantization to achieve impressive results in data compression and generation.  In VQ-VAE, the encoder maps the input data to a discrete latent representation using vector quantization, and the decoder reconstructs the input data from this discrete representation. Its ability to learn meaningful discrete representations has made it a popular choice for tasks that require high-quality, diverse sample generation~\cite{esser2021vqgan, razavi2019VQVAE2}. In our work, we build a learnable Disease Embedding Codebook for preserving disease latent space and mapping the disease-indicative features to it by vector quantization. However, we use only the discrete representations of the disease and the output of the encoder for the rest to reconstruct in decoder. 
\subsection{Generation Methods for Electrocardiograms}
\begin{figure*}
    \centering
    \setlength{\belowcaptionskip}{0mm}
    \includegraphics[width=0.95\linewidth]{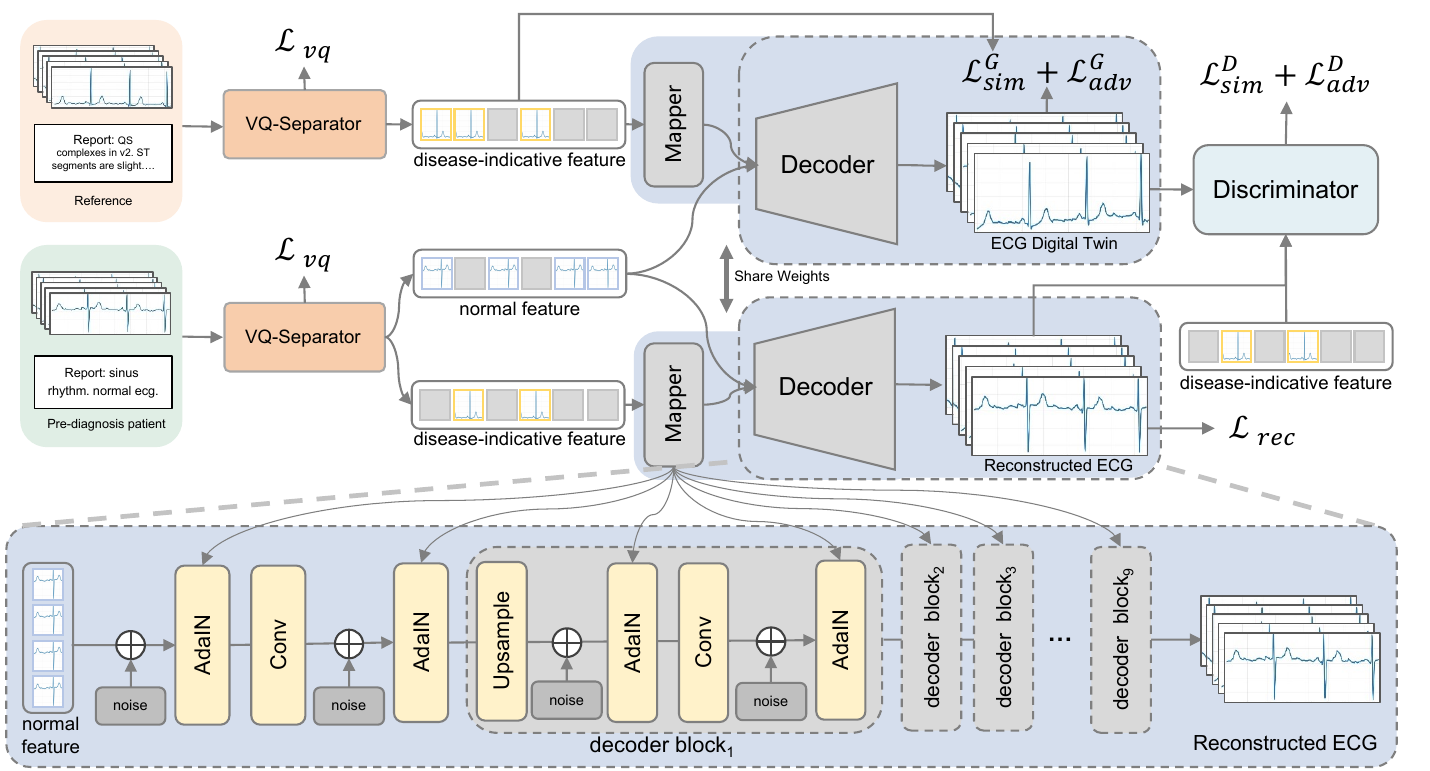}
    \caption{Overall framework of our LAVQ-Editor for ECG digital twin generation. Our method utilizes ECG-text pairs of the \textit{pre-diagnosis patient} and the \textit{reference patient}, merging personalized normal features and disease-indicative features to create ECG digital twins representing target heart disease symptoms.}
    \label{fig:main}
\end{figure*}

Electrocardiograms (ECGs) are a cornerstone in diagnosing heart diseases, yet their interpretation hinges on the discernment of clinical physicians, placing a considerable demand on their expertise and time. The imperative for more streamlined and effective automated systems for ECG analysis is driven by the need for rapid and precise heart health assessments. 
Efforts have been made to create automated ECG classifiers~\cite{baloglu2019classification,li2023mets,chen2020flow}, but they are often hampered by the scarcity of annotated ECG data and concerns over patient privacy. The synthesis of ECG signals emerges as a vital solution, enhancing the variety and volume of training samples while also mitigating privacy issues. Recent advancements, such as SimGAN~\cite{Golany2020simgan}, leverage Ordinary Differential Equations (ODEs) to capture cardiac dynamics and generate ECGs, whereas methodologies like Nef-Net~\cite{chen2021electrocardio} and ME-GAN~\cite{Chen2023megan} focus on the synthesis of multi-view ECG signals grounded in cardiac electrical activity. Besides, Chung~\etal~\cite{chung2023autotte} introduced Auto-TTE, an autoregressive generative model informed by clinical text reports. Golany ~\etal~\cite{golany2019pgans} generated personalized heartbeats by adding constraints on patient-relevant wave values. While existing studies often generate entirely new ECGs using disease-related information, our approach diverges by directly editing a patient's normal ECGs to incorporate specific disease features instead of constructing ECGs from noise. 

\section{Methods}
The manifestation of most heart diseases in ECG signals typically exhibits localized patterns, and such information is commonly included in ECG reports. Our objective is to generate ECGs with disease-specific characteristics while preserving the patient's individual traits according to the guidance provided in ECG reports. Refer to Figure~\ref{fig:main}, our model, LAVQ-Editor, comprises three components: a Vector Quantized Feature Separator (VQ-Separator) with a Disease Embedding Codebook, a generator $G$ consisting of a feature mapper and a decoder $G_{dec}$, and a discriminator $D$. Subsequently, we will introduce each of these three components individually.

\subsection{Vector Quantized Feature Separator}
\begin{figure}
    \centering
    \setlength{\belowcaptionskip}{0mm}
    \includegraphics[width=0.45\textwidth]{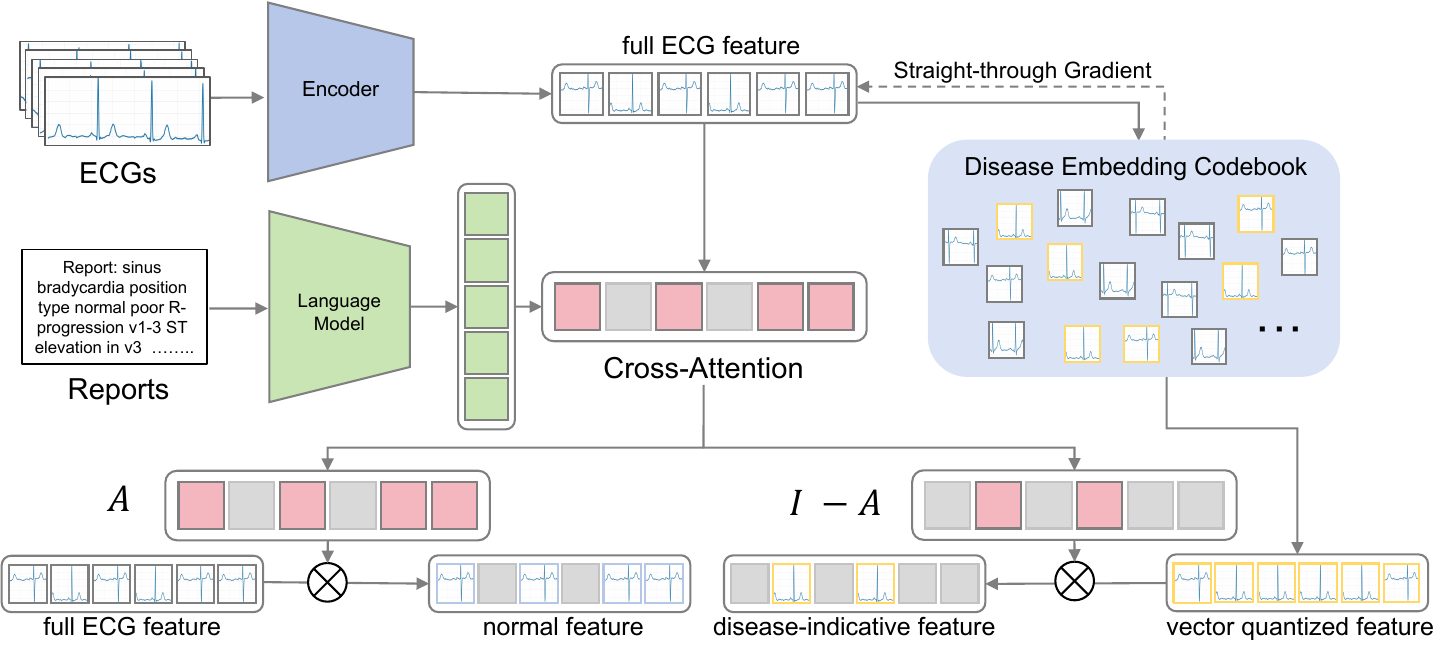}
    \caption{Overall Structure of Vector Quantized Feature Separator. Here, $A$ represents the cross-attention and $I$ represents an all-one matrix of the same size of $A$. ``ECGs'' represent the preprocessed input from 12-lead ECG data, and ``Reports'' refer to the texts of their corresponding diagnostic reports.}
    \label{fig:VQS}
\end{figure}

The role of the Vector Quantized Feature Separator is to separate the ECG features into normal features and disease-indicative features, guided by descriptive text. As shown in Figure~\ref{fig:VQS}, we first encode the ECGs with an encoder $G_{enc}$ based on 1D ResNet-34. This process can be represented as 
\begin{equation}
    f_{ecg} = G_{enc}(X_{ecg}), f_{ecg}\in \mathbb{R}^{C\times L_e}
\end{equation}

\noindent where $X_{ecg}$ is the input of normalized ECG signals. Here, $C$ and $L_e$ indicate the channel and the embedding dimensions, respectively. In practice, clinicians detect heart diseases by recognizing the disease-specific characteristics and properties of ECG signals, and document a rich source of information in the report. Inspired by~\cite{li2023mets}, we encode the ECG report texts with a pre-trained ClinicalBERT $G_{text}$~\cite{alsentzer2019clinalbert}, which is frozen in the training process. We further use a linear layer to compress $f_{text}$ to one dimension. Formally, the process can be formulated by
\begin{equation}
    f_{text} = \text{Linear}(G_{text}(X_{text})), f_{text} \in \mathbb{R}^{L_t}
\end{equation}
where $X_{text}$ is the descriptive text of the input ECG signals and $L_t$ indicate the embedding dimensions. 

Here we utilize the textual feature to guide the separation of features in ECG signals through cross-attention~\cite{hou2019crossatt}. Specifically, full ECG feature $f_{ecg}$ is refined into a query, while $f_{text}$ serves as both key and value. Then we define a criterion to determine whether a feature segment exhibits disease symptoms, simply implemented by a threshold $l$ for the cross-attention elements. Conversely, element values below this threshold are recognized as the normal features. This process can be defined as
\begin{equation}
    A = \mathds{1}[\text{MHCA}(f_{ecg}, f_{text})\geq l], 
\end{equation}
where $\mathds{1}[\cdot \geq l]$ binarizes the matrix elements based on whether they are larger than the threshold $l$, and MHCA means Multi-Head Cross-Attention module. The outcome $ A \in \mathbb{R}^{C\times L_e}$ is a binary mask indicating the segments presenting the disease symptoms. 

Utilizing the vector quantization technique, we compress and quantize the disease-indicative embedding within $f_{ecg}$ to remove personal information while accentuating disease symptoms. This process involves Disease Embedding Codebook (DEC), comprising a series of embedding vectors $v_k \in \mathbb{R}^{K \times L} $ for $ k \in \{1, 2, \ldots, K\} $, where $K$ represents the size of the embedding book. To isolate disease-indicative features, we select the nearest quantized discrete feature $Q(e)$ from the DEC. This procedure can be represented as: 
\begin{equation}
Q(e) := (argmin\parallel e_i - v_k\parallel_2^2 \ \mbox{for } i \mbox{ in } C). 
\end{equation}
Given that many ECG segments may not exhibit disease symptoms, we specifically choose segments corresponding to disease symptoms as the disease-indicative feature from $Q(e)$ calculated as $f_d = Q(e) * A $, where ``*'' denotes point-wise multiplication. Conversely, segments devoid of disease symptoms are selected as personal normal features from $e$, computed as $f_p = e * (I - A)$, with $I$ being an all-one matrix matching $A$'s shape. In this way, we effectively preserve normal features to retain individual information of \textit{pre-diagnosis patients}, while leveraging feature quantization to eliminate personal features from the \textit{reference patient} to maximize the retention of disease-related features.

Different from existing methods, our objective focuses on retaining the quantized latent disease embeddings within the DEC. Thus, there is no necessity to use the entirety of $Q(e)$. Instead, we selectively utilize only those segments identified as disease-indicative features. This is done with a loss function defined by: 
\begin{small}
\begin{equation}
    \mathcal{L}_{vq}=[\parallel (sg[e]-Q(e)) * A\parallel^2_2 + \parallel (sg[Q(e)]-e) * A\parallel^2_2] 
\end{equation}
\end{small}
\noindent where $sg[\cdot]$ means stop-gradient and ``*'' indicates the point-wise multiplication.
\subsection{Generator}
After obtaining disease-indicative features and personal normal features, we aim to fuse the target predicted disease-indicative features with personal normal features in the generator. In the training process, we use the \textit{reference patient}'s disease-indicative feature $f_d^t$ as the target disease, which is extracted in the same way as $f_d$. Before fusion, the disease-indicative features $f_d$ and $f_d^t$ are transformed into $\bar{f_d}$ and $\bar{f_d^t}$ by a mapper with eight sequential linear layers. Following~\cite{Karras2019stylegan,Chen2023megan}, we introduce adaptive instance normalization (AdaIN) to fuse $\bar{f_d^t}$ and $f_p$ in the decoder. We first add the learnable noise embedding to the personal normal features $f_p$ and then transform the disease-indicative features $\bar{f_d^t}$ to the control vectors, and finally perform AdaIN to obtain $\bar{f_p^0}$. Subsequently, we upsample $\bar{f_p^0}$ and further fuse it with $\bar{f_d^t}$ by nine decoder blocks with AdaIN and convolution layers, as shown in Figure~\ref{fig:main}. In each decoder block, the features are firstly upsampled with the nearest neighbor interpolation approach, and we inject $\bar{f_d^t}$ by two AdaIN operations. 

Unlike conventional generative models, our generator is tasked with producing not merely realistic ECGs but also with ensuring that the synthesized ECG digital twins accurately represent both the specific disease symptoms and the individual patient characteristics. To achieve this, we employ both reconstruction loss and disease similarity loss to optimize the generator. In order to accurately represent disease-indicative features, we try to ensure ECGs associated with the same symptoms to present similar features from DEC after processing by the VQ-Separator. Hence, we constrain the generated ECG digital twin to have target disease-indicative features by comparing the disease-indicative features separated from the reference ECG. Here we use the cosine similarity between the two to design the disease similarity loss, which is formally defined by:
\begin{gather}
    f_d^{G}, f_p^{G} = \text{VQ-Separator}(\hat{X}_{ecg}^t, X_{text}^t), \\
   \mathcal{L}_{sim}^G = 1 - \frac{f_d^G \cdot f_d^t}{\parallel f_d^G\parallel\parallel f_d^t\parallel}
\end{gather}
where $X_{text}^t$ represents the reports of the reference ECG.

Additionally, to ensure that a maximum amount of characteristics about the \textit{pre-diagnosis patient} is retained during the ECG editing process, we formulate a reconstruction loss. We use the disease-indicative feature $f_d$ of the ECG from the \textit{pre-diagnosis patient} instead of the target disease-indicative feature $f_d^t$ from the reference to reconstruct ECG by the generator. This process is defined as:
\begin{equation}
    \mathcal{L}_{rec} = \parallel \hat{X}_{ecg}^{rec} - X_{ecg} \parallel^2
\end{equation}
\noindent where $\hat{X}_{ecg}^{rec}$ means the reconstructed ECG by the generator. In line with established generative adversarial networks, we also employ an adversarial loss to optimize the generator, by:
\begin{equation}
    \mathcal{L}_{adv}^G = \mathbb{E}_{{z} \sim {p_z(z)}}(log(1+exp(-D(G(z)))))
\end{equation}
where $p_z$ represents the distribution of the \textit{pre-diagnosis patients}' ECGs. Thus, the total optimization objective of the generator can be represented as:
\begin{equation}
    \mathcal{L}^G = \mathcal{L}_{adv}^G + \mathcal{L}_{rec} + \mathcal{L}_{sim}^G + \mathcal{L}_{vq}.
\end{equation}

\subsection{Discriminator}
Our discriminator architecture, featuring ten blocks and two heads, is designed for authenticity verification. In the architecture, each block contains two convolution layers with a 3-sized kernel, followed by a ReLU activation function for non-linearity, and a Gaussian blur layer with a kernel size of 2 for feature smoothing.

The output head, responsible for verifying the authenticity of the ECG, incorporates a layer of Standard Deviation~\cite{Karras2019stylegan}, a convolution layer and two linear layers that finally determines the authenticity of the ECG. The optimization objective for the discriminator also includes the conventional discriminator loss, by: 
\begin{equation}
\begin{aligned}
       \mathcal{L}_{adv}^D &= \mathbb{E}_{{x} \sim {p_{data}(x)}}(log(1+exp(-D(x)))) +\\
    &\mathbb{E}_{z \sim {p_g(z)}}(log(1+exp(D(z)))) 
\end{aligned}
\end{equation}

\noindent where $p_{data}$ represents the distribution of the real ECGs and $p_g$ represents the distribution of the generated ECGs.

In our study, generating realistic ECGs that precisely capture the symptoms of specific heart diseases is also paramount. To achieve this, we designed the \textit{disease-indicative feature extraction head} that has the similar structure with the output head, but with a key distinction in the output size of the final linear layer. To equip the discriminator with the ability to assess disease-indicative features in ECGs, we optimize it by maximizing the similarity between $f_d$ extracted from the \textit{pre-diagnosis patient} and $f_d^D$ extracted from the reconstructed ECG by the \textit{disease-indicative feature extraction head}, as:
\begin{equation}
    \mathcal{L}_{sim}^D = 1 - \frac{f_d^D \cdot f_d}{\parallel f_d^D\parallel\parallel f_d\parallel}.
\end{equation}

The optimization objective of the discriminator is:
\begin{equation}
    \mathcal{L}^D = \mathcal{L}_{adv}^D + \mathcal{L}_{sim}^D.
\end{equation}
\section{Experiments}
Our evaluation of the proposed method is organized on three aspects: effectiveness, fidelity, and privacy security. To assess effectiveness, we incorporate our prospective learning method into a data augmentation model for heart disease detection.  We then introduce two specialized evaluation metrics to measure the fidelity of our synthesized ECG digital twins, drawing parallels with established benchmarks in image generation. Additionally, we compute the ``membership inference risk'' metric to assess the privacy security of the generated results. 
Our study includes comparisons with conventional generative methods condition GAN~\cite{mirza2014conditiongan}, VQ-VAE2~\cite{razavi2019VQVAE2}, WGAN~\cite{Arjovsky2017wgan}, StyleGAN~\cite{Karras2019stylegan},  and personalized generative method PGANs~\cite{golany2019pgans}.

\subsection{Implementation Details}
In our experiments, we utilize the PTB-XL dataset, a comprehensive and publicly accessible collection of electrocardiograms, which contains 21,837 clinical 12-lead ECGs from 18,885 individuals. Accompanying each ECG is an extensive report and a set of diagnostic labels. These labels are categorized into 5 overarching superclasses and an additional 24 subclasses. Our research strictly focuses on the superclasses, using them as the basis for all analyses presented in this study. Given the importance of clear disease differentiation, our experiments are conducted exclusively with single-label data. To ensure a robust test of our model's capabilities, we selected 291 patients who have both normal and disease ECG records for testing, while allocating the remaining patients' data for training and validation purposes. 

For raw ECG data, preprocessing is performed to align the positions. The peak detection algorithm from the Python package NeuroKit2 is utilized to extract the indices of the R peaks in the ECG. The segments from 100 points before the first R peak to 100 points after the seventh R peak are extracted. These segments are then scaled proportionally to a fixed length of 4096 points for the following experiments.

For the training process of LAVQ-Editor, we employ the Adam optimization algorithm, with a learning rate set at 0.00001 and a weight decay of 0.001. The model undergo a rigorous training regime spanning 500 epochs and is processed with a batch size of 128. The threshold $l$ is set as 0.5 in the following experiments and we provide ablation study of the threshold in the supplements. All subsequent experiments are executed using the PyTorch 1.9 on an NVIDIA GeForce RTX-3090 GPU. We benchmark our method against the existing generative models tailored to accommodate the data and maintain consistency in the training configuration to ensure fairness in comparison. 
\subsection{Personalized heart disease detection}
\begin{table*}[htpb]
\centering
\begin{tabular}{l|ccc|ccc}
\toprule
  \multicolumn{1}{c|}{\multirow{2}{*}{\centering Method}}        & \multicolumn{3}{c|}{Patient Wise} & \multicolumn{3}{c}{ECG Wise} \\ \cline{2-7}
                               & Accuracy($\uparrow$)& F1-score($\uparrow$)   & ATE($\uparrow$)     & Accuracy($\uparrow$)& F1-score($\uparrow$) & AUROC($\uparrow$)    \\ \hline
ResNet-34 w/o generator        & 69.13\%  & 0.6358     & 0.41\%  & 66.41\%  & 0.5259   & 0.7604         \\
ResNet-34+CGAN                 & 70.13\%  & 0.6309     & 1.09\%  & 67.97\%  & 0.5160   & 0.8332         \\
ResNet-34+WGAN                 & 70.94\%  & 0.6370     & 0.19\%  & 68.75\%  & 0.5289   & 0.8450         \\
ResNet-34+StyleGAN             & 70.62\%  & 0.6409     & 0.63\%  & 68.36\%  & 0.4702   & \textbf{0.8637}         \\
ResNet-34+VQ-VAE2              & 69.54\%  &0.6124      & 2.14\%  &69.53\%   & 0.5699   & 0.8193    \\
ResNet-34+PGANs                 & 69.67\%  & 0.6296     & 1.87\%  & 68.36\%  & 0.4748   & 0.8539         \\
ResNet-34+LAVQ-Editor(Ours) & \textbf{72.21\%} & \textbf{0.6717}  & \textbf{5.13\%} & \textbf{70.70\%}  & \textbf{0.5988}   & 0.8450     \\ \bottomrule
GoogleNet w/o generator&67.19\% & 0.6088 & 0.48\%  & 66.02\%  & \textbf{0.5858}  & 0.8474  \\
GoogleNet+CGAN        & 67.90\% & 0.6198 & 1.34\% & 65.63\% & 0.5157 & 0.8026   \\
GoogleNet+StyleGAN    & 67.60\% & 0.6136 & 1.28\% & 65.63\% & 0.4867 & 0.8011   \\
GoogleNet+VQ-VAE2     & 69.73\% & 0.6371 & 1.63\%  & 66.80\% & 0.5142 & 0.7862   \\
GoogleNet+PGANs       & 69.31\% & 0.6234 & 0.25\%  & 66.80\% & 0.5689 & 0.8578   \\
GoogleNet+LAVQ-Editor(Ours) & \textbf{70.73\%} & \textbf{0.6500}  & \textbf{1.94\%} & \textbf{67.97\%}  & 0.5476   & \textbf{0.8605}   \\ \bottomrule
SVM w/o generator& 38.00\% & 0.2693 &0.00\% & 38.95\%  & 0.2026  & 0.4886  \\
SVM+CGAN        & 40.37\%  & 0.2004 & 2.29\% & 40.41\% & 0.1765 & 0.4607   \\
SVM+StyleGAN    & 38.63\% & 0.2755 & 1.02\% & 39.24\% & 0.1936 & 0.4898   \\
SVM+VQ-VAE2     & 39.13\% & 0.2746 & 2.12\%  & 39.24\% & 0.1882 & 0.4831   \\
SVM+PGANs       & 39.60\% & 0.2930 & 1.28\%  & 40.12\% & 0.1786 & 0.4822   \\
SVM+LAVQ-Editor(Ours) & \textbf{40.63\%} & \textbf{0.2956}  & \textbf{3.77\%} & \textbf{41.28\%}  & \textbf{0.2065}   & \textbf{0.5248}     \\ \bottomrule
\end{tabular}
\vskip -0.6 em
\caption{Evaluation of the Classification Performance of Several Base Models Trained on Augmented Training Sets Synthesized via Various Techniques. The best performances are marked in \textbf{bold}.}
\label{tab:cls}
\end{table*}

To verify the utility of our method, we test whether there is a benefit of prospective learning for personalized heart disease detection models. We use 1D ResNet-34, 1D GoogleNet and SVM as the base model and compare the performance of the base model enhanced with and without the results of different generative models. In this study, we randomly divide the patients in the test set into two even groups for a comprehensive evaluation of our prospective learning methodology. The first group, designed as the experimental group, is subjected to our prospective learning approach, providing prospective cognition into the heart disease detection model. The second group is assigned as the control group, serving as a baseline against which the effectiveness of our methodology could be compared. For every normal ECG in the experimental group of the test set, we generate 50 ECGs with diseases and add them to the training set. We optimize the base model by the Adam optimizer with a learning rate of 0.001 and weight decay of 0.001, and train 256 epochs with a batch size of 64. 

\paragraph{Evaluation at the patient level.}To evaluate the performance of our methodology for personalized heart disease detection, we implement metrics of personalized accuracy, F1-score, and Average Treatment Effect (ATE). Personalized accuracy and F1-score represent the aggregate of individual patient accuracies and F1-scores, respectively. The ATE metric is particularly insightful, quantifying the differential in expected accuracy outcomes when a specific treatment is applied across a patient cohort, as opposed to when it is withheld. Mathematically, ATE is defined as:
\begin{equation}
    ATE = \mathbb{E}[\mathrm{Accuracy}(W=1)-\mathrm{Accuracy}(W=0)],
\end{equation}
\noindent where $W=1$ denotes the application of generated related ECGs and $W=0$ signifies its absence. The higher the Average Treatment Effect (ATE) score, the more the method improves the average effectiveness of the diagnosis of a \textit{pre-diagnosis patient}'s disease.

As shown in Table~\ref{tab:cls}, our prospective learning method culminates in the most commendable patient-wise accuracy and F1-score among all tested configurations. Regarding the ATE metric, our method demonstrates substantial benefits, indicating that the personalized ECG digital twins can significantly bolster disease diagnosis. It is worth noting that the increase in accuracy is not as marked as the enhancement seen in ATE. This disparity is largely attributable to sample forgetting~\cite{toneva2018sampleforget} observed in the control group, which occurred after the augmentation of generated results by our method. This leads to a reduction in accuracy among the control group, thereby widening the performance gap between it and the experimental group. In clinical application, our primary objective is to improve the diagnostic outcomes for \textit{pre-diagnosis patients}. Therefore, the decrease within the control group is of lesser consequence.

\paragraph{Evaluation at the ECG level.} We use traditional metrics evaluated on ECG levels for heart disease detection models, i.e., classification accuracy, F1-score, and Area Under the Receiver Operating Characteristic (AUROC). In the right-hand section of Table~\ref{tab:cls}, it is obvious that our method achieves the best ECG-wise accuracy and F1-score among all evaluated methods. While our AUROC score isn't the highest, it still signifies a robust ability to differentiate between classes. Notably, even though models like PGANs and StyleGAN register higher AUROC scores than ours, their F1-scores are significantly lower. As reflected in the F1-score, our method balances both precision and recall. Therefore, these results collectively demonstrate that our method markedly enhances the overall efficacy of ECG analysis in the context of heart disease detection.

\paragraph{Impacts of the quantity of generated ECG digital twins.}Here we conducted an experiment to validate the impact of the quantity of generated ECG digital twins on the results. We trained a 1D ResNet model with varying amounts of digital twins and examined the performance discrepancies. As shown in Table~\ref{tab:GvR}, it is evident that increasing the quantity of digital twins improves accuracy at both the patient and ECG levels. However, on ATE, it appears that having more generated digital twins does not yield better results. This could be attributed to duplicated cases introducing redundant information, thus compromising the model's effectiveness in capturing disease-related patterns.
\begin{table}[htpb]
\centering
\setlength{\abovecaptionskip}{1mm}
\setlength{\belowcaptionskip}{-4mm}
\scalebox{0.7}{
\begin{tabular}{cc|ccc|ccc}
\toprule
\multicolumn{2}{c|}{Data Num}          & \multicolumn{3}{c|}{Patient Wise}                                  & \multicolumn{3}{c}{ECG Wise}                                       \\ \hline
\multicolumn{1}{c|}{Real}  & Generated & \multicolumn{1}{c|}{Acc($\uparrow$)}    & \multicolumn{1}{c|}{F1($\uparrow$)}     & ATE($\uparrow$)    & \multicolumn{1}{c|}{Acc($\uparrow$)}    & \multicolumn{1}{c|}{F1($\uparrow$)}     & AUROC($\uparrow$)  \\ \hline
\multicolumn{1}{c|}{15553} & 8600      & \multicolumn{1}{c|}{72.21\%} & \multicolumn{1}{c|}{0.6717} & \textbf{5.13\%} & \multicolumn{1}{c|}{70.70\%} & \multicolumn{1}{c|}{\textbf{0.5988}} & 0.8450 \\
\multicolumn{1}{c|}{15553} & 17200     & \multicolumn{1}{c|}{73.55\%} & \multicolumn{1}{c|}{0.6690} & 2.05\% & \multicolumn{1}{c|}{70.70\%} & \multicolumn{1}{c|}{0.5919} & 0.8555 \\
\multicolumn{1}{c|}{15553} & 34400     & \multicolumn{1}{c|}{\textbf{73.81\%}} & \multicolumn{1}{c|}{\textbf{0.6759}} & 2.33\% & \multicolumn{1}{c|}{\textbf{71.48\%}} & \multicolumn{1}{c|}{0.5392} & \textbf{0.8605} \\ \bottomrule
\end{tabular}
}
\caption{The impact of the quantity of generated ECG digital twins on performance. }
\label{tab:GvR}
\end{table}

\subsection{Generated ECG fidelity evaluation}
To evaluate the authenticity of the ECGs, we adapt metrics from image generation tasks, specifically the Fréchet Inception Distance (FID)~\cite{heusel2017fid} and the Inception Score (IS)~\cite{Salimans2016is}, traditionally used for assessing the quality of generated images. Diverging from the conventional approach, we utilize a pre-trained 1D ResNet-101 model instead of the Inception V3~\cite{Szegedy2016inceptionv3} as the backbone for feature extraction from ECG data. Consequently, we introduce the ``Fréchet ResNet Distance'' (FRD) and ``ResNet Score'' (RS) as our novel metrics for quantifying the fidelity of ECG synthesis. Lower FRD scores are indicative of a closer resemblance between the distribution of generated and real ECGs, signifying a higher level of authenticity in the synthesized data. As outlined in Table~\ref{tab:fid}, our method distinguishes itself among various generative approaches by achieving the lowest FRD score, markedly surpassing its counterparts. In terms of the RS, where higher values are indicative of better performance, our method registers an impressive RS of 9.9771, significantly exceeding those achieved by other methods. This underscores our method's effectiveness in closely mirroring the actual ECG distribution.

We further adopt Precision and Recall proposed by Kynkaanniemi~\etal~\cite{kynkaanniemi2019improved} and the performances are shown in Table~\ref{tab:fid}. In addition, we also report the comprehensive F1-score balancing the Precision and Recall. Evidently, our method exhibits significant advantages in both F1-Score and Recall, suggesting that ECGs generated by our approach ensure sample quality and provide extensive coverage. While CGAN demonstrates superior Precision, the markedly low Recall suggests that it generates homogeneous ECG replicating training data.

\begin{table}[t]
\centering
\resizebox{0.48\textwidth}{!}{
\begin{tabular}{l|ccccc}
\toprule
         \multicolumn{1}{c|}{Method}       & FRD($\downarrow$)   & RS($\uparrow$) & Precision($\uparrow$)  & Recall($\uparrow$) & F1-Score($\uparrow$)\\ \hline
Real       & 0.3014 & 16.8195  & - & - & - \\ \hdashline
CGAN       &4.2389    &1.7866 &\textbf{0.9968}&0.0688 &0.1287 \\
WGAN       &3.9171    &1.8853 &0.9000    &0.2250 &0.3600\\
StyleGAN   &5.4443 & 1.8412 &0.8969 & 0.0781  &0.1437 \\
VQ-VAE2    &4.2539  & 3.5462 &0.9094  & 0.4187 &0.6240\\
PGANs      &3.8477    &2.0853 &0.7781    &0.4750 &0.5899\\
Ours       & \textbf{1.6286} & \textbf{9.9771} & 0.8719 & \textbf{0.6438} &\textbf{0.7407}       \\ \bottomrule
\end{tabular}
}
\vskip -0.8 em
\caption{Evaluation of the fidelity of ECGs generated by diverse generative models. The best performances are marked in \textbf{bold}.}
\label{tab:fid}
\end{table}

\begin{figure}[t]
\centering  
\includegraphics[width=\linewidth]{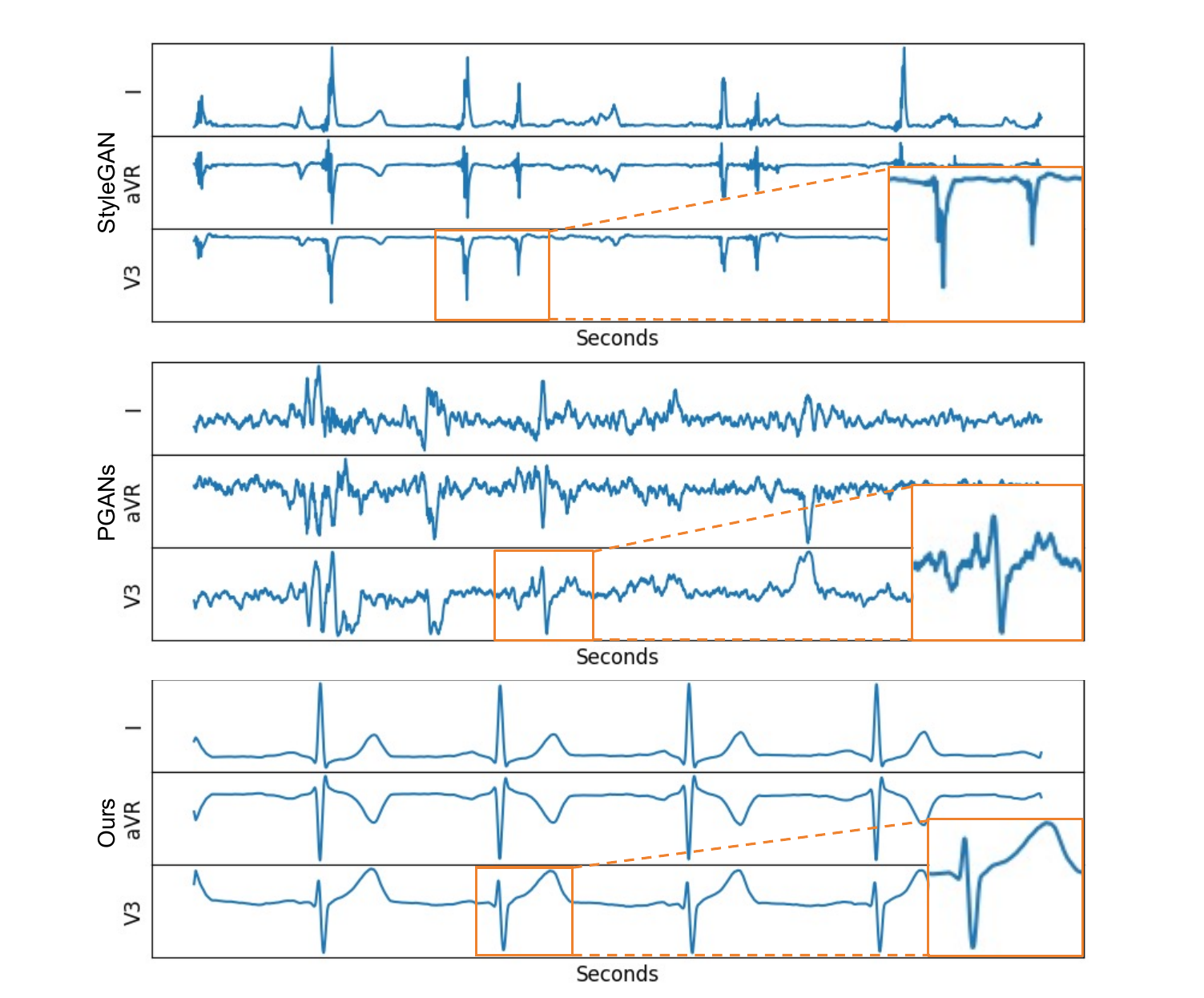}
\setlength{\abovecaptionskip}{-0.3cm}
\setlength{\belowcaptionskip}{-3mm}
\caption{Visualization of ECG generated of STTC (ST/T Change) category by different models. For clarity in comparison, we only use the electrocardiogram of three leads: I, aVR, and V3.}
\label{fig:ecg}
\end{figure}

To straightforward assess the fidelity of our model, we conduct a direct analysis of ECGs generated by various models, as shown in Figure~\ref{fig:ecg}. Overall, StyleGAN-generated ECGs exhibit regular rhythms but suffer from excessive noise and lack realism. ECGs produced by PGANs designed for personalized generation, struggle with identification in wave peak. Our method, however, not only captures the fundamental characteristics of a heartbeat but also significantly reduces noise, thus enhancing the realism of the ECGs. Focusing on the STTC category disease symptoms, marked by the orange rectangles, we observe that while StyleGAN's results prominently feature T-wave inversion, it's difficult to determine whether this is due to noise or an actual anomaly. The outputs of PGANs, are also marred by considerable noise across three leads, blurring the line between noise and real T-wave anomalies. In sharp contrast, our approach clearly demonstrates a significant ST segment elevation in the V3 lead, a definitive indication of a pathological feature. This stark distinction indicates that our method generates results with great advantages in terms of fidelity as well as accuracy of disease characterization. In addition, we will provide more examples in the supplements.
\subsection{Privacy security performance}\label{sec:privacy}

\begin{figure}
\centering  
\setlength{\abovecaptionskip}{-1mm}
\setlength{\belowcaptionskip}{-3mm}
\includegraphics[width=0.9\linewidth]{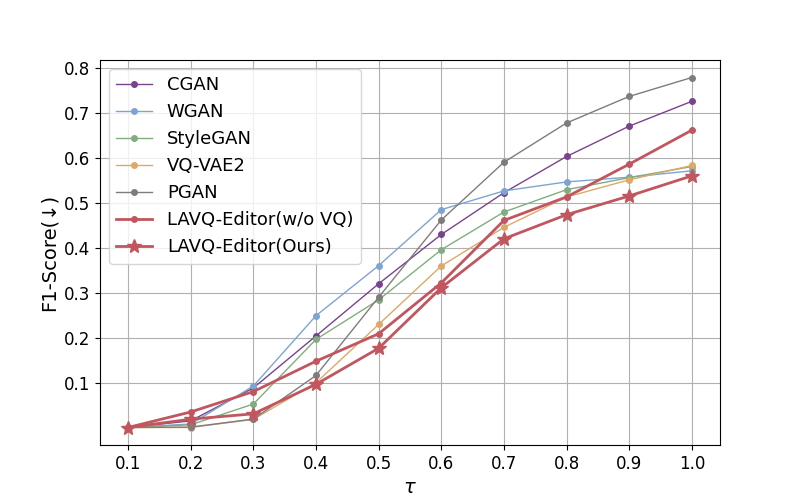}
\caption{The F1-score for the membership inference risk. The horizontal coordinate indicates $\tau$, and the vertical coordinate indicates the F1-score. The higher the F1-score, the higher the risk. }
\label{fig:mem}
\end{figure}

In clinical practice and machine learning development, safeguarding patient privacy is also important. We rigorously assess the synthetic ECG digital twins for any risk of presence disclosure, ensuring that the information within these digital twins is non-sensitive and cannot be linked back to the original patient data. To assess the risk of presence disclosure, a scenario where an adversary could deduce the inclusion of specific samples in the model’s training set from synthetic ECGs, we employ the concept of membership inference risk~\cite{shokri2017membership}. This type of disclosure becomes a potential issue when an attacker, possessing a subset of target ECGs from the training set, attempts to analyze synthetic ECGs to confirm the presence of these target samples in the training data. Given a distance threshold $\epsilon$, the adversary claims that a target ECG is in the real training set if there exists at least one ECG with a distance less than the threshold. In our experiments, we set the threshold $\epsilon$ based on the mean value of the distance between all ECGs, which can be expressed as $\epsilon = \tau\times mean$. To compute the distance between ECGs, we first extract the ECG features using 1D ResNet-101 pretrained on training set and then compute the Euclidean distance between the features. Finally, we use F1-score as membership inference risk. The higher the F1-score, the higher the risk. As shown in Figure~\ref{fig:mem}, Our method demonstrates a lower risk of presence disclosure compared to the other methods across the most of thresholds. Notably, although our model uses samples from the training set when generating ECGs, our risk is still lower than other models that do not use samples from the training set, indicating that our model is effective in preventing presence disclosure. Additionally, the line LAVQ-Editor(w/o VQ) shows the performance of the model without VQ-Separator. When compared with the full LAVQ-Editor, there is a heightened risk of presence disclosure. This suggests that the VQ-Separator effectively separates the personal normal feature of the \textit{reference patient} from their disease-indicative feature, thereby enhancing privacy security.

\subsection{Personal information retention performance}
\begin{figure}
\centering  
\setlength{\belowcaptionskip}{-3mm}
\includegraphics[width=0.85\linewidth]{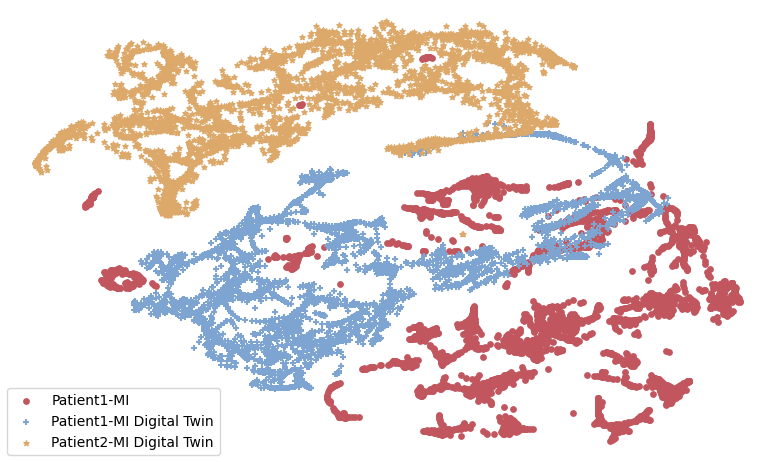}
\caption{t-SNE visualization demonstrating the distinction between ECG digital twins created using the same reference but for different pre-diagnosis patients. }
\label{fig:tsne}
\end{figure}

To further demonstrate the retention of personal information in the digital twins, we utilized t-SNE to visualize differences between ECG digital twins created using the same reference but for different pre-diagnosis patients. As depicted in Figure~\ref{fig:tsne}, ``Patient1-MI'' represents the reference ECG with Myocardial Infarction derived from Patient1, while ``Patient1-MI Digital Twin'' refers to an ECG generated for Patient1's normal ECG using ``Patient1-MI'' ECG as a reference. Similarly, ``Patient2-MI Digital Twin'' indicates that Patient2's normal ECG is used as the pre-diagnosis patient's ECG. Figure~\ref{fig:tsne} clearly shows greater overlap between the ``Patient1-MI'' and ``Patient1-MI Digital Twin'' distributions, whereas ``Patient2-MI Digital Twin'' is distinctly separated from both, highlighting that the ``Patient1-MI Digital Twin'' successfully retains personal information of Patient1. 
Remarkably, this section prioritizes retaining personal information in pre-diagnosis patients' ECGs. In contrast, Section~\ref{sec:privacy} is focused on removing identifiable information from the references. The marked similarity between ``Patient1-MI'' and ``Patient1-MI Digital Twin'' and their pronounced distinction from ``Patient2-MI Digital Twin'' exemplify this strategy effectively. 

\subsection{Ablation Study}
In addition, we also perform ablation study to validate the rationale of our framework.  By employing Dynamic Time Warping (DTW)~\cite{Keogh2001DerivativeDTW} which is acclaimed for its precision in aligning varying temporal sequences, we evaluate the distance between generated ECG digital twin and the real ECGs from \textit{pre-diagnosis patients} with the same disease which indicates similarity to the \textit{pre-diagnosis patients}. Given that ECGs naturally differ in length and rhythm, DTW is particularly adept at adapting these time series to closely align their patterns, enhancing the reliability of the comparison. 
As detailed in Table~\ref{tab:abl}, our ablation study results suggest that vector quantization is pivotal for the retention of personalized disease characteristics in the heart disease detection model. Without it, the model may produce ECGs that closely mirror the original but fail to enhance the detection model effectively. The implementation of both generator's and discriminator's disease similarity losses is instrumental in imbuing the generated ECGs with disease traits. However, this narrowly focused approach comes at the expense of individualized patient information, as reflected by the diminished ATE and high DTW observed in the third line of Table~\ref{tab:abl}. As shown in the second and third lines, reconstruction loss ensures the synthesized ECG digital twins retain patient-specific characteristics and are not disproportionately influenced by disease features, which is crucial for achieving lower DTW scores. In addition, the absence of either similarity loss in the generator or discriminator results in unstable adversarial learning dynamics, preventing the model from reaching its optimal performance.

\begin{table}
\centering
\begin{tabular}{cccc|cc}
\toprule
VQ &$\mathcal{L}_{sim}^G$ & $\mathcal{L}_{sim}^D$ & $\mathcal{L}_{rec}$ & DTW ($\times10^3$, $\downarrow$) & ATE ($\uparrow$)\\ \hline
          &\checkmark   & \checkmark    & \checkmark  & 6.664  &1.24\%\\
\checkmark&\checkmark   & \checkmark    &             & 44.096 &2.76\% \\
\checkmark&             &               &             & 12.264 &1.98\% \\
\checkmark&             &               & \checkmark  & 5.772  &1.52\%      \\ 
\checkmark&             & \checkmark    & \checkmark  & 5.519  &1.35\% \\ 
\checkmark&\checkmark   &               & \checkmark  & 5.674  &2.55\% \\ 
\checkmark&\checkmark   & \checkmark    & \checkmark  & \textbf{5.322}&\textbf{5.13\%} \\ \bottomrule
\end{tabular}
\caption{Evaluation of the significance of different components of the proposed framework. The best performances are marked in \textbf{bold}. VQ: Vector Quantized Feature Separator; DTW: Dynamic Time Warping; ATE: Average Treatment Effect.}\label{tab:abl}
\end{table}

\section{Conclusions}

In this paper, we introduce a novel \textit{prospective learning} framework for creating personalized ECG digital twins that simulate the heart conditions of diagnosed individuals. 
For the purpose of realistic ECG generation, we propose a location-aware ECG digital twin generation model based on vector quantization, termed LAVQ-Editor. The model skillfully segregates and edits normal and disease-indicative features in ECGs, preserving individual characteristics while injecting target disease information. 
This enhances understanding of personalized heart diseases prospectively while safeguarding privacy in model development. To the best of our knowledge, our LAVQ-Editor is the first framework for personalized ECG digital twin generation in contrast to previous work that only generated data at population level. The versatility of our approach extends beyond diagnostics, holding potential for application in ethically-conducted scientific research, such as clinical trial simulation.

\appendix



\section*{Acknowledgments}
This research was supported by Zhejiang Provincial Natural Science Foundation of China (LY23F020019), National Natural Science Foundation of China (62106218 \& 82200558), Excellent Young Scientists Fund (82122036), Basic and Applied Basic Research Foundation of Guangdong Province (2022A1515110722), and Guangdong Provincial Medical Science \& Technology Research Fund Project (A2023011).

\bibliographystyle{named}
\bibliography{s_bib}
\section*{Appendix A: Pseudocode of Training}
\begin{algorithm}
\begin{algorithmic}
\STATE {Initialize VQ-Separator, Generator $G$, and Discriminator $D$}
\FOR{each training epoch}
    \FOR{each batch in the dataset}
        \STATE $X_{ecg} \gets$ pre-diagnosis ECG, $X_{text} \gets$ the report of ECG
        \STATE $X_{ecg}^t \gets$ reference ECG, $X_{text}^t \gets$ the report of reference ECG
        \STATE \textbf{Optimizer Discriminator $D$:}
        \STATE $f_d, f_p \gets  \text{VQ-Separator}({X}_{ecg}, X_{text})$
        \STATE $f_d^t, f_p^t \gets  \text{VQ-Separator}({X}_{ecg}^t, X_{text}^t)$
        \STATE $\hat{X_{ecg}^t} \gets  G_{dec}(f_p, f_d^t)$
        \STATE $\hat{X}_{ecg}^{rec} \gets  G_{dec}(f_p, f_d)$
        \STATE $f_d^{G}, f_p^{G} \gets  \text{VQ-Separator}(\hat{X}_{ecg}^t, X_{text}^t)$
        \STATE $\hat{f_d}, D_{real} \gets  D(X_{ecg})$
        \STATE $\hat{f_d^{rec}}, D_{fake} \gets  D(\hat{X}_{ecg}^{rec})$
        \STATE Calculate losses:
        \STATE $L_{adv}^D \gets \log(D_{real}) + \log(1 - D_{fake})$ 
        \STATE $L_{sim}^D \gets \mathcal{L}_{sim}^D = 1 - \frac{f_d^D \cdot f_d}{\parallel f_d^D\parallel\parallel f_d\parallel}$
        \STATE Update $D$ by minimizing $ L_{sim}^{D} + L_{adv}^{D}$
        \STATE \textbf{Optimizer Generator $G$:}
        \STATE $f_d, f_p, L_{vq} \gets  \text{VQ-Separator}({X}_{ecg}, X_{text})$
        \STATE $f_d^t, f_p^t \gets  \text{VQ-Separator}({X}_{ecg}^t, X_{text}^t)$
        \STATE $\hat{X_{ecg}^t} \gets  G_{dec}(f_p, f_d^t)$
        \STATE $\hat{X_{ecg}^{rec}} \gets  G_{dec}(f_p, f_d)$
        \STATE $f_d^{G}, f_p^{G} \gets  \text{VQ-Separator}(\hat{X}_{ecg}^t, X_{text}^t)$
        \STATE $\hat{f_d^{rec}}, D_{fake} \gets  D(\hat{X}_{ecg}^{rec})$
        \STATE Calculate losses:
        \STATE $L_{recon} \gets ||X_{ecg} - \hat{X}_{ecg}^{rec}||^2 $
        \STATE $L_{adv} \gets \log(1 - D_{fake})$ 
        \STATE Update $G$ by minimizing $L_{recon} + L_{vq} + L_{adv}^G + L_{sim}^G$
    \ENDFOR
    \STATE Evaluate model performance on validation set
\ENDFOR
\end{algorithmic}
\caption{Training Process of the LAVQ-Editor}
\end{algorithm}

\section*{Appendix B: Ablation Study of threshold $l$}

\begin{figure}[!h]
\centering  
\includegraphics[width=0.95\linewidth]{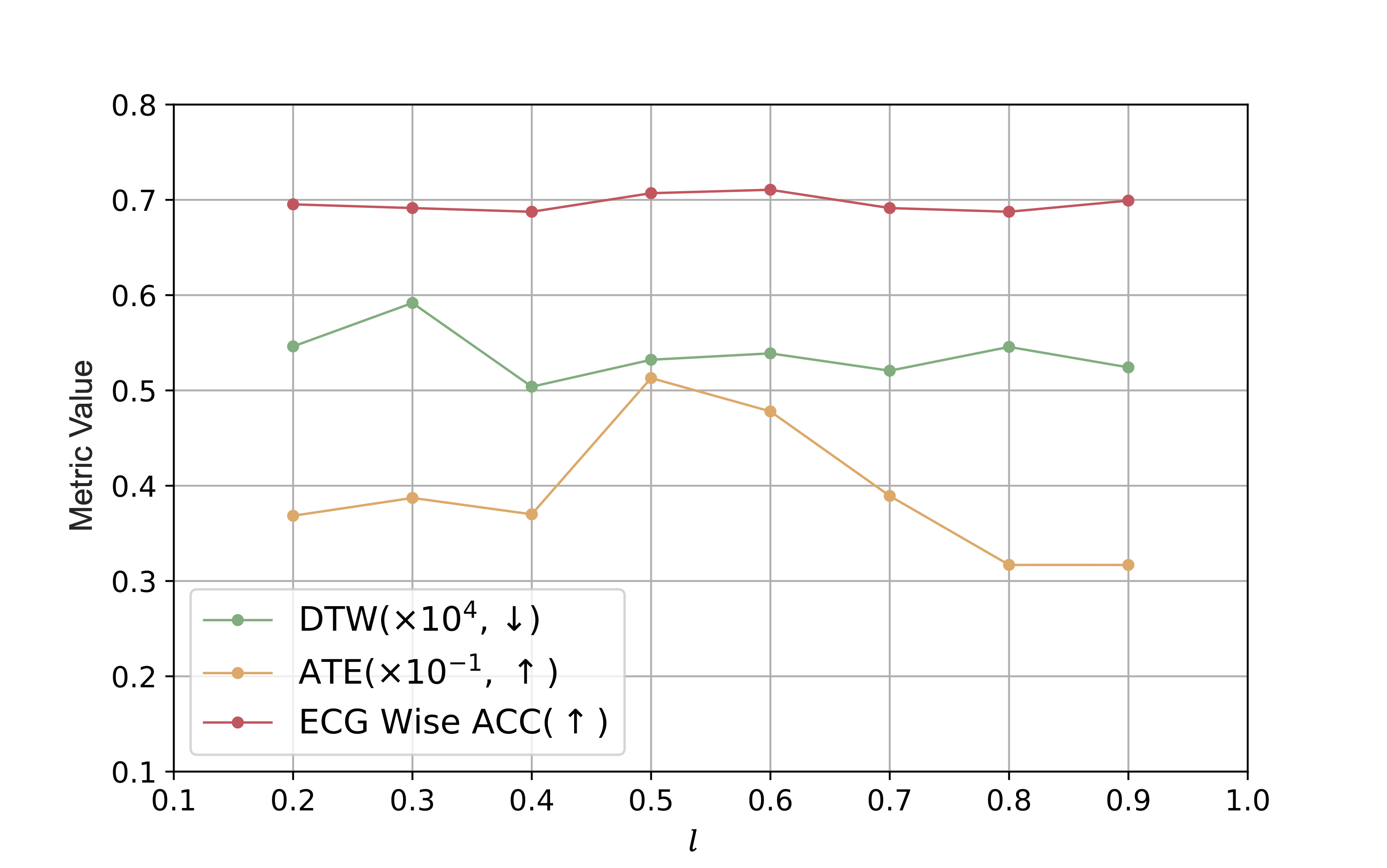}
\caption{Model performance with different values of the threshold, $l$. DTW: Dynamic Time Warping; ATE: Average Treatment Effect; ECG Wise ACC: ECG Wise Accuracy. }
\label{fig:abl-l}
\end{figure}

The Vector Quantized Feature Separator incorporates textual features to navigate the dissection of features within ECG signals via a cross-attention mechanism. Within VQ-Separator, we establish a simple criterion for identifying disease symptoms in feature segments, hinging on a threshold $l$ for the cross-attention elements. Segments with values above this threshold $l$ are flagged as disease-indicative segments, while those falling below are classified as normal segments.

To assess the influence of the threshold $l$ on the authenticity of the synthesized ECGs, we conduct experiments across a spectrum of threshold values. As depicted in Figure~\ref{fig:abl-l}, the Dynamic Time Warping (DTW)—scaled by $10^4$ for display purposes—remains relatively stable under varying thresholds, maintaining values below $0.6 \times 10^4$. The ECG wise accuracy shows minimal fluctuation, consistently averaging around 0.7. In contrast, the Average Treatment Effect (ATE), represented in Figure~\ref{fig:abl-l} after being scaled by $10^{-1}$, exhibits more pronounced variability. Nonetheless, it sustains a level above $0.3 \times 10^{-1}$, indicating a robust consistency in the validity of the results.

\section*{Appendix C: Extra Visualization Results}
\begin{figure}
\centering  
\includegraphics[width=\linewidth]{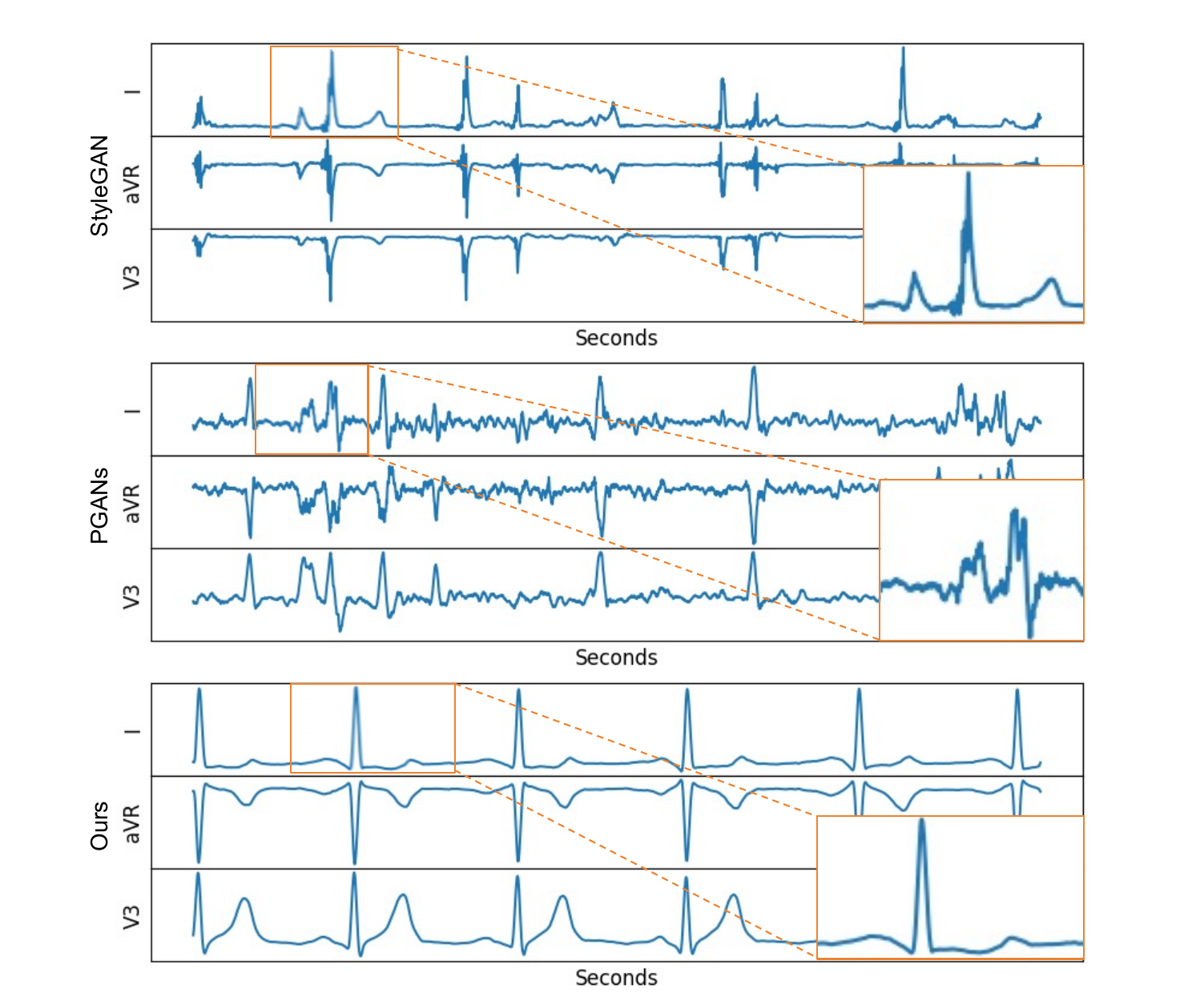}
\caption{Visualization of generated ECGs with the Hypertrophy (HYP) heart disease patterns by different models. For better viewing, we only display three leads of ECGs: I, aVR, and V3.}
\label{fig:HYP}
\end{figure}

\begin{figure}[h]
\begin{minipage}[t]{.5\textwidth}
    \centering  
    \includegraphics[width=\linewidth]{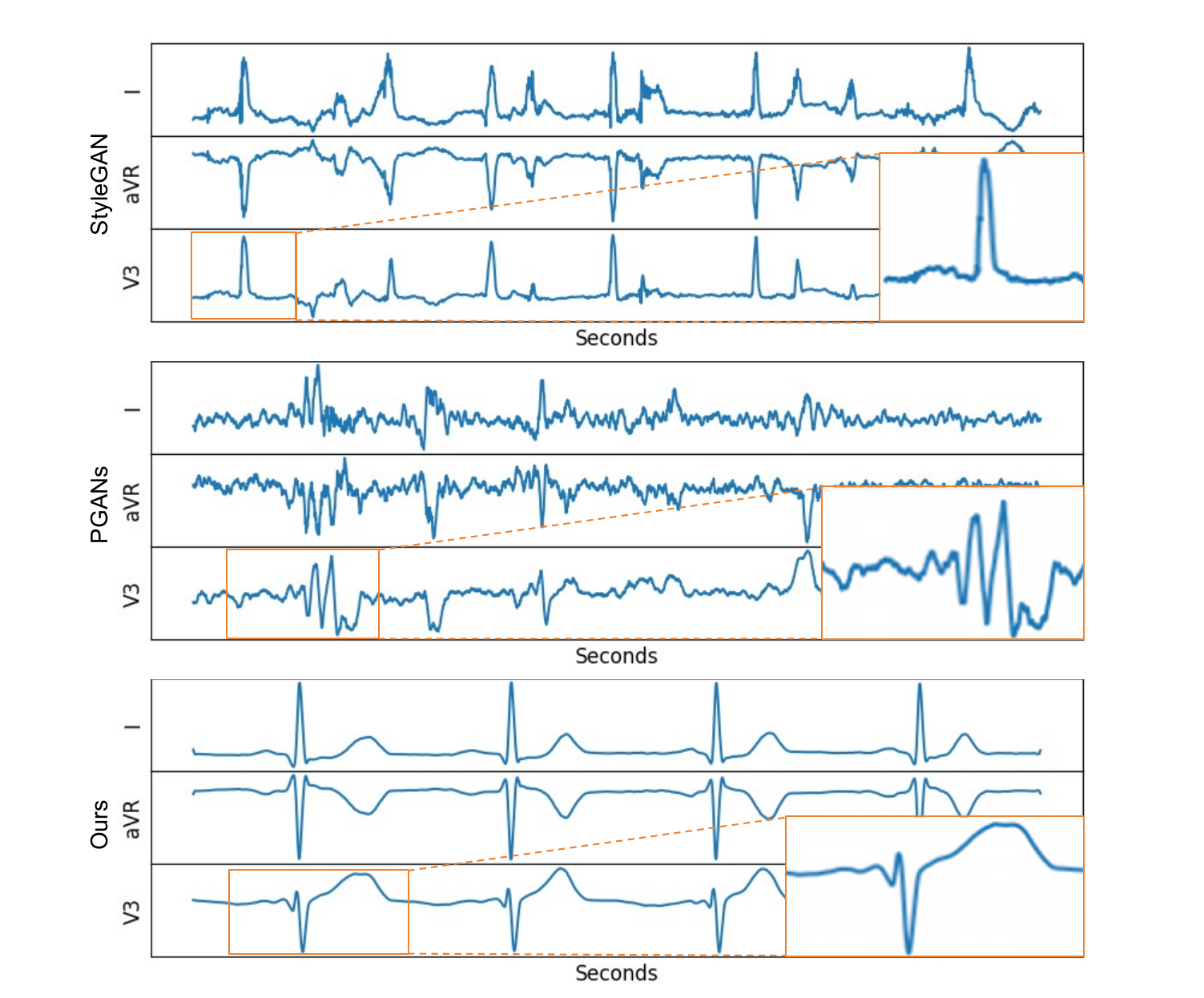}
    \caption{Visualization of generated ECGs with the Myocardial Infarction (MI) heart disease patterns by different models. For better viewing, we only display three leads of ECGs: I, aVR, and V3.}
    \label{fig:MI}
\end{minipage}
\\
\begin{minipage}[t]{.5\textwidth}
    \centering  
    \includegraphics[width=\linewidth]{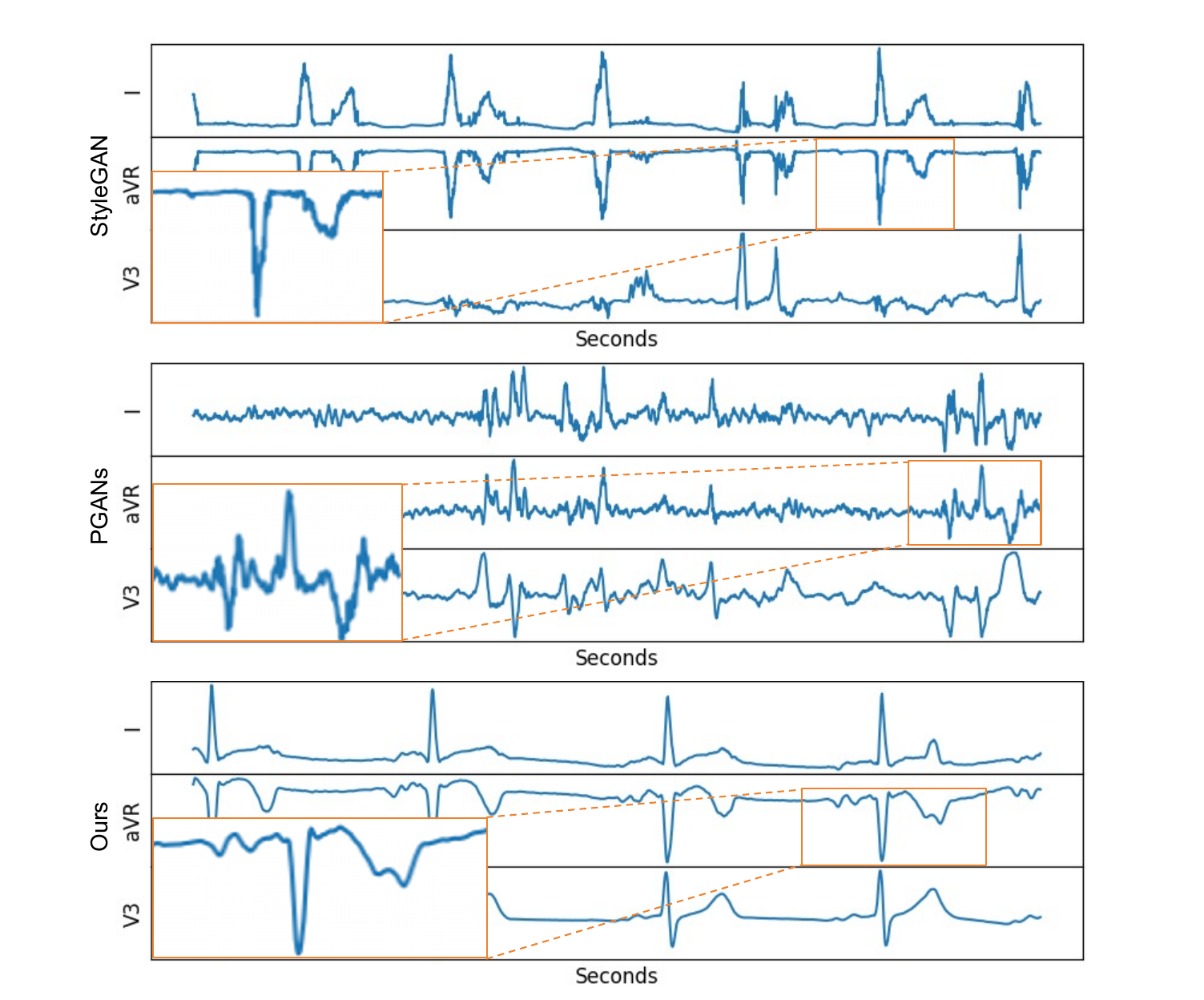}
    \caption{Visualization of generated ECGs with the Conduction Disturbance (CD) heart disease patterns by different models. For better viewing, we only display three leads of ECGs: I, aVR, and V3.}
    \label{fig:CD}
\end{minipage}
\end{figure}

Referring to Figures ~\ref{fig:HYP}, ~\ref{fig:MI}, and ~\ref{fig:CD}, we display and intuitively inspect the generated ECGs with heart diseases: Hypertrophy, Myocardial Infarction, and Conduction Disturbance, by various models. Our findings reveal:

\noindent (i) ECGs generated by StyleGAN are regularly rhythmic but are plagued by excessive noise, diminishing their authenticity. As a result, distinguishing genuine abnormalities from noise becomes a challenge, reducing their clinical utility. 

\noindent (ii) PGANs, designed for personalized generation, still struggles with accurate wave depiction, hindering the identification of disease symptoms.

\noindent (iii) Our method successfully generates correct heartbeat features and markedly reduces noise, boosting ECG realism. Furthermore, our approach distinctly reveals clear pathological signs. As indicated by orange rectangles in Figure ~\ref{fig:HYP}, we observe T-wave flattening in lead I, indicative of potential hypertrophic cardiomyopathy. Figure ~\ref{fig:MI} display pronounced QRS abnormalities in lead V3, where the T wave is notably higher than the R and S waves. Figure ~\ref{fig:CD} shows instances of the P wave failing to progress to the ventricle, potentially indicative of ventricular or junctional escape rhythms. 

\begin{figure}[!b]
    \centering
\setlength{\abovecaptionskip}{0.5mm}
\setlength{\belowcaptionskip}{-3mm}
    \includegraphics[width=0.8\linewidth]{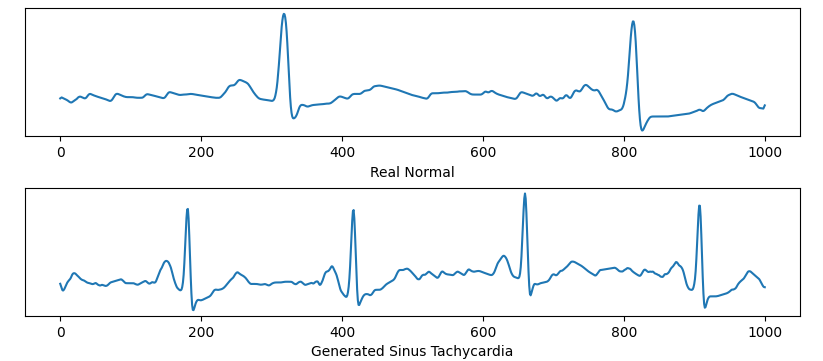}
    \caption{Example of ECG described as ``sinus tachycardia apart from rate, normal ecg." generated from the real normal ECG.}
    \label{fig:sinustachycardia}
\end{figure}

For timing problems, as shown in Figure~\ref{fig:sinustachycardia}, our model can scale the signals based on the statistical RR interval length after morphological editing.

\end{document}